\documentclass[runningheads]{llncs}
\usepackage{eccv}
\usepackage{eccvabbrv}
\usepackage{graphicx}
\usepackage{booktabs}
\usepackage[accsupp]{axessibility} 
\usepackage{amsmath}
\usepackage{multirow}
\usepackage{colortbl}
\newcommand{\myparagraph}[1]{\vspace{1mm}\noindent{\bf #1}}

\usepackage{hyperref}
\usepackage{orcidlink}

\begin{document}

\title{Fourier Decomposition for Explicit Representation of Colored Point Cloud Attributes} 
\titlerunning{Fourier-based Colored Point Cloud Encoding}

\author{
Donghyun Kim\inst{1, 2} \and
Chanyoung Kim\inst{1, 2}\and
Hyunah Ko\inst{1} \and
Seong Jae Hwang\inst{1}
}
\authorrunning{D. Kim et al.}

\institute{
Yonsei University, Seoul, Republic of Korea \\
\email{\{danny0103, chanyoung, kha9867, seongjae\}@yonsei.ac.kr}
\and
Emory University, Atlanta GA, USA \\
\email{\{donghyun.kim, chanyoung.kim\}@emory.edu}
}

\maketitle

\begin{center}
    \centering
    \includegraphics[width=\textwidth]{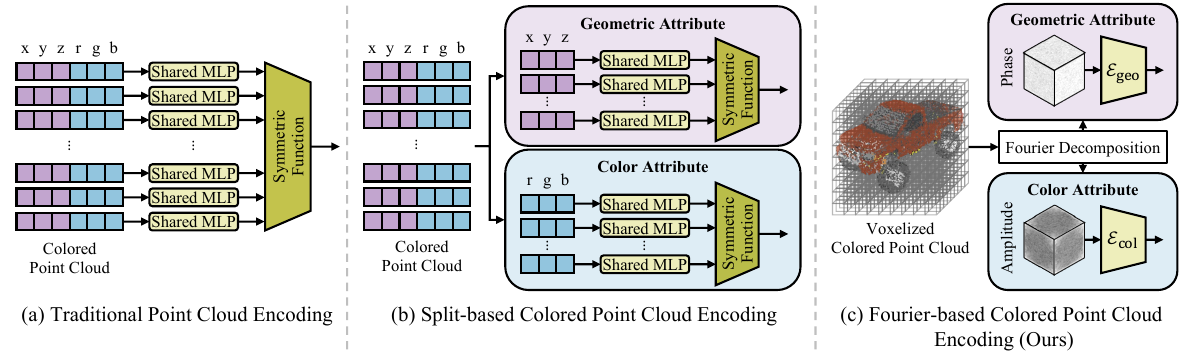}
    \captionof{figure}{Comparison of existing point cloud encoding approaches. We present \textit{a Fourier-based colored point cloud encoding method} that explicitly leverages amplitude and phase to represent the color and geometric attributes of the point cloud. This encoding enables effective processing across various point cloud tasks, including classification, segmentation, and style transfer.}
    \label{fig:overview_2col}
\end{center}

\begin{abstract}
While 3D point clouds are widely used in vision applications, their irregular and sparse nature make them challenging to handle. 
In response, numerous encoding approaches have been proposed to capture the rich semantic information of point clouds. 
Yet, a critical limitation persists: a lack of consideration for colored point clouds, which serve as more expressive 3D representations encompassing both color and geometry.
While existing methods handle color and geometry separately on a per-point basis, this leads to a limited receptive field and restricted ability to capture relationships across multiple points.
To address this, we pioneer a colored point cloud encoding methodology that leverages 3D Fourier decomposition to disentangle color and geometric features while extending the receptive field through spectral-domain operations.
Our analysis confirms that our approach effectively separates feature components, where the amplitude uniquely captures color attributes and the phase encodes geometric structure, thereby enabling independent learning and utilization of both attributes.
We validate our colored point cloud encoding approach on classification, segmentation, and style transfer tasks, achieving state-of-the-art results on the DensePoint dataset. 
All the attached source code will be made publicly available.
\end{abstract}

\section{Introduction}
\label{sec:introduction}

Point clouds have gained increasing attention with the advancement of fields such as autonomous driving~\cite{auto1, auto2} and augmented reality~\cite{augment1, augment2}, due to their effectiveness in representing 3D information.
In response, various deep learning tasks leveraging point clouds (\textit{e.g.}, classification~\cite{pointmixer, pointasnl, paconv}, segmentation ~\cite{randla, stratified, gacnet}, and upsampling~\cite{platypus, grad_pu, repkpu}) have been extensively explored. 
However, unlike 2D images that consist of discrete grids, point clouds represent sparse sets of points in continuous space.
This irregularity has introduced challenges in processing point clouds.

Typically, a point cloud is represented as an \( N \times c_0 \) matrix, where \( N \) is the number of points and \( c_0 \) denotes the input channels. 
Each point is initially represented by Cartesian coordinates (\( x, y, z \)), but \( c_0 \) can be extended to include additional attributes such as RGB color values (\( r, g, b \)) or surface normals (\( n_x, n_y, n_z \)). 
Building on this representation, early deep learning methods~\cite{pointnet, pointnet++} employ shared multi-layer perceptrons (MLPs) to extract features from the \( c_0 \) channels of each point (\cref{fig:overview_2col}(a)).
While this approach ensures permutation invariance and mitigates certain irregularities, it does not account for the unique characteristics of \textit{colored point clouds}, which contain rich information for both geometric and color attributes within the input channels.
This leads to a failure in distinguishing these independent attributes, treating them as a single, entangled feature.
Hence, it often failed to sufficiently capture the semantic information of the colored point cloud.

To facilitate the learning of colored point cloud features, prior works~\cite{psnet, colorpcr} have independently handled geometric and color attributes through split-based point cloud encoding (\cref{fig:overview_2col}(b)), which partitions the colored point cloud ($N\times6$) into two separate groups: one containing the XYZ coordinates ($N\times3$) and the other containing the RGB values ($N\times3$). 
Each group is then fed into a separate feature extractor. 
However, as feature extractors used in this approach extract features on a per-point basis, it inherently suffers from a limited receptive field, making it challenging to capture context-rich features that emerge when considering neighboring points.
Therefore, the development of a proper encoding method
that not only handles the two attributes independently but also incorporates neighboring points for feature extraction remains a crucial research direction in the field of colored point cloud processing.

In response to this direction, we pioneer a colored point cloud encoding method that utilizes amplitude and phase readily obtained through a newly devised 3D Fourier decomposition (\cref{fig:overview_2col}(c)).
We have discovered that amplitude preserves the point cloud’s color information, whereas phase encodes its geometric attributes. 
Since both components reside in the spectral domain, operations applied to them influence the entire point cloud globally~\cite{ffc_evidence}, resulting in a large receptive field. 
Leveraging this property, we employ amplitude and phase as \textit{explicit representations}, facilitating independent learning and effective utilization of the point cloud’s color and geometric features, making them a suitable encoding method for colored point clouds.

To demonstrate the practical utility of our findings, we apply them to three key tasks: (i) classification, (ii) segmentation, and (iii) style transfer. 
For classification and segmentation, our approach achieves state-of-the-art performance on the DensePoint~\cite{densepoint} dataset. 
In style transfer, we utilize the amplitude within an optimization-based framework to modify the style of a point cloud, achieving more effective style transfer compared to existing method.

\myparagraph{Contributions.} 
In this work, we introduce a Fourier-based colored point cloud encoding that employs the components obtained through Fourier decomposition as explicit representations. 
Our main contributions are:
\begin{itemize}
    \item We are the first to analyze colored point clouds using 3D Fourier decomposition, uncovering a key insight: amplitude uniquely preserves detailed color attributes, while phase represents the intrinsic geometric structure.

    \item We present a Fourier-based colored point cloud encoding framework that disentangles color and geometric features from the colored point cloud. This framework enables the independent utilization of color and geometric attributes, with achieving a large receptive field. 

    \item We demonstrate the utility of our findings through its application to classification, segmentation, and style transfer. The proposed methods perform favorably against other methods on the DensePoint and Semantic3D datasets.
\end{itemize}
\section{Related Work}
\label{sec:related_works}

\myparagraph{Point Cloud Encoding Approaches.}
Early approaches for processing 3D point clouds made significant efforts to effectively handle their irregular nature (\textit{i.e.}, unordered sets of points).
Various methods have been proposed, including multi-view projections~\cite{projec2_multidetect, rotationnet}, which project point clouds onto multiple image planes, and voxel-based networks that quantize point clouds into voxel grids~\cite{octnet, rel_voxel1, rel_voxel2}. However, these approaches often fail to capture fine geometric details~\cite{projec1_multicnn, projec3_veh}.
To overcome these limitations, pointwise-processing methods~\cite{pointnet, pointnet++} address permutation invariance by employing shared MLPs and a symmetric function (\textit{e.g.}, max pooling).
PointNet~\cite{pointnet} processes each point independently to capture global features, while PointNet++~\cite{pointnet++} introduces set abstraction, which utilizes hierarchical grouping, to capture both local and global features.
Transformer-based methods~\cite{point_transformer, point_transformer_v2, point_transformer_v3}
employ self-attention with pointwise operations for efficient local and global feature extraction.
While these studies have effectively addressed the geometric attributes of 3D point clouds, they have largely overlooked the color and style-related attributes, which are as crucial as geometry in the case of colored point clouds. 
Our work fills this gap by analyzing colored point clouds and introducing explicit representations containing geometry and style attributes.

\myparagraph{Fourier-based Analysis on 2D Image.}
In the 2D image domain, the Fourier Transform has been widely applied to tasks such as domain adaptation and generalization, leveraging the insight that the amplitude captures domain-specific attributes.
FACT~\cite{fact} enhances domain generalization by interpolating amplitude spectra, 
whereas FDA~\cite{fda} transfers style from a target image by swapping low-frequency amplitude components.
Similarly, TF-Cal~\cite{tf_cal} calibrates the amplitude at test time to handle variations in style.
In image restoration~\cite{falcon, LaMa}, methods utilizing the Fourier Transform to address image degradation have also emerged.
Fourmer~\cite{fourmer} exploited the disentanglement of degradation elements through Fourier Transform, confirming that its phase preserves the underlying semantic structure while its amplitude encodes domain-specific appearance. 
Despite these successes in 2D images, where data is organized as uniform grids of discrete pixels, applying Fourier Transform to 3D point clouds is challenging due to their sparse and unstructured nature.
Although studies~\cite{gft1, gft2, gft3} have applied the Graph Fourier Transform to 3D point clouds, they rely on graphs constructed from point clouds rather than the raw point clouds in their entirety, thus lacking analysis of the pure attributes that can be derived from the complete structure of the point cloud in 3D space.
Our work addresses this challenge by being the first to implement Fourier decomposition for colored point clouds, demonstrating that the resulting amplitude and phase can be employed as explicit representations that capture semantic information in point clouds.

\section{Fourier-based Colored Point Cloud Encoding}
\label{sec:fourier_decomposition}

In this section, we present the newly devised Fourier decomposition and reconstruction tailored for point cloud, which are fundamental to our encoding approach, and provide an in-depth analysis of the resulting amplitude and phase.
We first detail the adaptation of the Fourier Transform for irregular point cloud structures (\cref{subsec:technical_details}).
Next, we analyze the specific information encoded in each component (\cref{subsec:experimental_analysis}).
A theoretical rationale for the inherent property of each component is provided in the supplementary.

\subsection{Fourier Decomposition and Reconstruction}
\label{subsec:technical_details}
\begin{figure}[t]
    \centering
    \includegraphics[width=\linewidth]{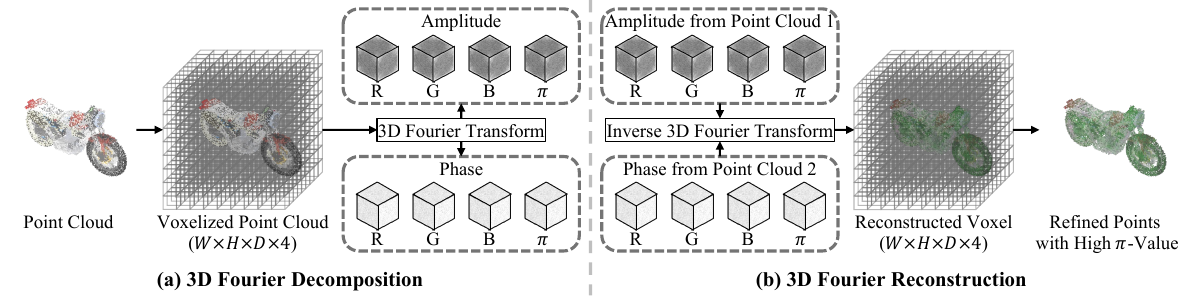}
    \caption{
    Detailed process of 3D Fourier decomposition and reconstruction.
    (a) The point cloud is voxelized and decomposed via Fourier Transform into amplitude and phase. An additional channel $\pi$ represents the probability of a point existing in each voxel.
    (b) The inverse Fourier Transform reconstructs voxel, removing low-\(\pi\) voxels to reduce amplitude-phase misalignment noise.
    }
    \label{fig:fourier_decomposition_implementation_details}
\end{figure}

To perform Fourier decomposition and Fourier reconstruction (\cref{fig:fourier_decomposition_implementation_details}) of point cloud data, which sparsely resides in a continuous space, we propose an implementation method that accounts for the sparse distribution of point clouds.

\myparagraph{3D Fourier Decomposition.}
The Discrete Fourier Transform is typically designed to operate on data arranged in fixed arrays such as 2D images. To extend the Fourier Transform to point cloud, which is irregular data representation, we map the point cloud $\mathbf{P}$ onto a structured 3D grid through voxelization.
The boundaries of the voxel grid are defined by computing the smallest (\(x_{\min}, y_{\min}, z_{\min}\)) and largest (\(x_{\max}, y_{\max}, z_{\max}\)) values along the x, y, and z axes and define the voxel grid's extent accordingly. 
To preserve as many points as possible by assigning the minimum number of points per voxel, we set a sufficiently small voxel size $v$.
Using the voxel size and the boundaries, we compute the shape of the voxel grid $(W, H, D)$ as
\begin{equation}
\small
    W, H, D = \frac{x_{\max} - x_{\min}}{v}, \frac{y_{\max} - y_{\min}}{v}, \frac{z_{\max} - z_{\min}}{v}.
\end{equation}

Considering the sparse nature of point clouds (\textit{i.e.}, not all voxels contain points), we introduce \textit{a probabilistic occupancy channel $\pi$}, to represent the probability of a point being present in each voxel. 
For voxels containing points, the \(\pi\) value is assigned as 1, while for voxels without points, the \(\pi\) value is set to 0. This channel is stacked with the RGB channels, forming a voxel grid of size $W \times H \times D \times 4$, where the four channels correspond to R, G, B, and $\pi$. 
We then apply a 3D Discrete Fourier Transform to the voxel data $V$ to obtain the Fourier coefficients $\hat{V}$ as follows:
\begin{equation}
\small
    \hat{V}(k, l, m) = \sum^{W-1}_{x=0} \sum^{H-1}_{y=0} \sum^{D-1}_{z=0} V(x, y, z) e^{-2\pi i\left(\frac{kx}{W} + \frac{ly}{H} + \frac{mz}{D}\right)}.
\end{equation}
From these Fourier coefficients, we derive the amplitude $\mathcal{A} = |\hat{V}|$ and phase $\mathcal{P} = \arg(\hat{V})$, thereby constructing a frequency-domain representation of the point cloud data.

\myparagraph{3D Fourier Reconstruction.}
To reconstruct a point cloud from the four channels (R, G, B, $\pi$) of amplitude and phase obtained through 3D Fourier decomposition, we apply the inverse Discrete Fourier Transform to convert these components into a single voxel representation on a 3D grid.  
If the amplitude-phase pair originates from the same point cloud without any modifications, the voxelized representation of the original point cloud, generated during Fourier decomposition, is accurately restored.
However, when any alterations are made to the amplitude or phase pair, their alignment is disrupted, leading to a reconstructed voxel data that deviates from an ideal structure and introduces significant noise.  
To eliminate this noise (\textit{i.e.}, outliers), we leverage the $\pi$ channel to selectively determine the final points for reconstruction.
Similar to the RGB channels, the $\pi$ channel is processed through both the Fourier Transform and its inverse, encoding the probability of a point's presence within each voxel.
If a voxel's $\pi$ value falls below a predefined threshold, indicating a low probability of containing a point, it is treated as empty space in the continuous domain and we generate no points in the corresponding spatial region.
Conversely, if the $\pi$ value exceeds the threshold, the voxel is determined to contain a valid point, and the corresponding RGB values are retained for reconstruction in the continuous space.

\begin{figure*}[t]
    \centering
    \includegraphics[width=\linewidth]{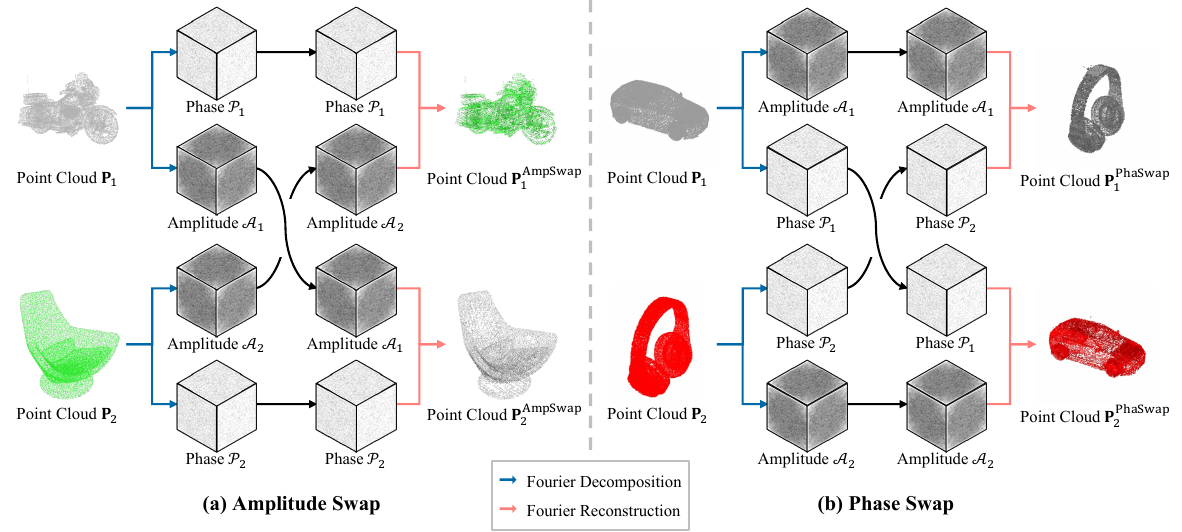}
    \caption{
    An experimental investigation into Fourier decomposition in point clouds reveals the distinct roles of amplitude and phase. Performing Fourier decomposition separately on two point clouds and exchanging their amplitude components before reconstruction results in a transfer of color attributes. In contrast, exchanging phase components alters the geometric structure, indicating that amplitude primarily encodes color information, whereas phase captures the underlying spatial arrangement.
    }
    \label{fig:fourier_decomposition_swap}
\end{figure*}

\subsection{Intrinsic Analysis of Amplitude and Phase}
\label{subsec:experimental_analysis}
To verify the type of information encoded in the amplitude and phase obtained through Fourier decomposition, we swap either the amplitude or phase between two different point clouds, as illustrated in \cref{fig:fourier_decomposition_swap}.
Given two distinct point clouds, $\mathbf{P}_1$ and $\mathbf{P}_2$, we perform Fourier decomposition $\mathcal{F}$ to obtain their respective amplitude $\mathcal{A}$ and phase $\mathcal{P}$ components as
\begin{equation}
\small
    \mathcal{A}_i, \mathcal{P}_i = \mathcal{F}(\mathbf{P}_i), \quad i =1, 2.
    \label{eqation:decom}
\end{equation}
\myparagraph{Amplitude Swap.}
With the given amplitude and phase in Eq.~\eqref{eqation:decom}, we swap the amplitude to demonstrate that amplitude encodes the color attributes of the point cloud. 
Specifically, we replace the amplitude of one point cloud with that of another while keeping the phase unchanged, then reconstruct the point clouds using Fourier reconstruction $\mathcal{F}^{-1}$ as  
\begin{equation}
\small
    \mathbf{P}^\text{AmpSwap}_i = \mathcal{F}^{-1}(\mathcal{A}_{3-i}, \mathcal{P}_i),  \quad i =1, 2.
\end{equation}
As illustrated in \cref{fig:fourier_decomposition_swap}(a), this swapping process results in an exchange of color-related attributes between the two point clouds, despite their structural information remaining unchanged (see supplementary for more examples). 
This demonstrates that the amplitude component extracted through Fourier decomposition primarily encodes color attributes in the point cloud representation.

\myparagraph{Phase Swap.}
We investigate the inherent property of the phase component by swapping it between two point clouds while keeping the amplitude unchanged. The point clouds are then reconstructed using $\mathcal{F}^{-1}$ as  
\begin{equation}
\small
    \mathbf{P}^\text{PhaSwap}_i = \mathcal{F}^{-1}(\mathcal{A}_{i}, \mathcal{P}_{3-i}),  \quad i =1, 2.
\end{equation}
As shown in \cref{fig:fourier_decomposition_swap}(b), this operation leads to an exchange of geometry-related attributes while preserving the original color characteristics. In contrast to the amplitude-swapping experiment, where only color information was transformed, swapping the phase alters the structural properties of the point cloud. 
The phase component obtained via Fourier decomposition mainly represents geometric attributes, as confirmed by this.
Additional results on amplitude and phase swapping can be found in the supplementary material.

\section{Downstream Applications}
\label{sec:applications}

In this section, we introduce downstream applications of our encoding method, which leverages the intrinsic properties of amplitude and phase (introduced in Sec.~\ref{sec:fourier_decomposition}).
Specifically, we outline its potential use in three representative tasks: (\romannumeral 1) classification (Sec.~\ref{subsec:classification}), (\romannumeral 2) segmentation (Sec.~\ref{subsec:segmentation}), and (\romannumeral 3) style transfer (Sec.~\ref{subsec:point_cloud_style_transfer}).

\subsection{Point Cloud Classification}
\label{subsec:classification}
\begin{figure}[t]
    \centering
    \includegraphics[width=0.7\linewidth]{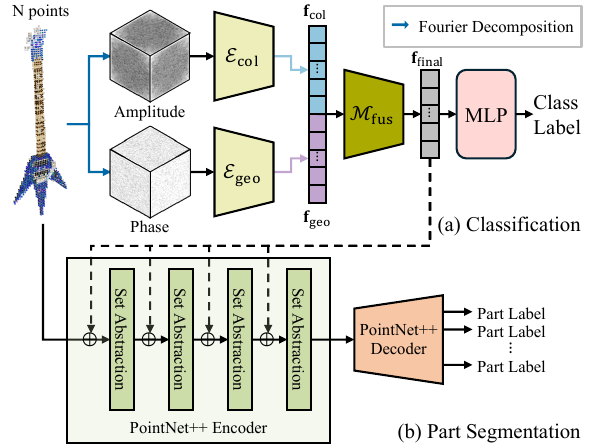}
    \caption{
    An overview of our model using Fourier-based point encoding.  
    Input points are decomposed into amplitude and phase through Fourier decomposition, then separately processed by a color encoder $\mathcal{E}_\text{col}$ capturing color attributes and a geometry encoder $\mathcal{E}_\text{geo}$ capturing geometric attributes. 
    The resulting feature vectors are concatenated and processed through a fusion module $\mathcal{M}_\text{fus}$ to generate the final feature representation $\mathbf{f}_\text{final}$, which is then utilized in both classification and segmentation.
    }
    \vspace{-10pt}
    \label{fig:feature_extractor}
\end{figure}

To demonstrate the effectiveness of our colored point cloud encoding, we design a classification model for colored point clouds based on the proposed Fourier-based encoding approach, as shown in \cref{fig:feature_extractor}(a).
Notably, we adopt a simple model architecture, integrating the Fourier-based encoding with a few learnable layers, highlighting that strong performance can be achieved without complex designs.
First, we apply Fourier decomposition into input point cloud, separating it into amplitude $\mathcal{A}$ and phase $\mathcal{P}$ components. 
We encode these components into feature vectors that capture color and geometric attributes. Specifically, a color encoder \(\mathcal{E}_\text{col}\) extracts a feature vector from the amplitude, defined as \(\mathbf{f}_\text{col} = \mathcal{E}_\text{col}(\mathcal{A}) \in \mathbb{R}^{D_\mathcal{A}}\), while a geometry encoder \(\mathcal{E}_\text{geo}\) derives a feature vector from the phase, given by \(\mathbf{f}_\text{geo} = \mathcal{E}_\text{geo}(\mathcal{P}) \in \mathbb{R}^{D_\mathcal{P}}\). 
Here, \( D_\mathcal{A} \) and \( D_\mathcal{P} \) denote the dimensions of each feature.
These two feature vectors are concatenated and processed through a fusion module \(\mathcal{M}_\text{fus}\) to generate the final feature representation \(\mathbf{f}_\text{final}\). The entire classification model is then trained using cross-entropy loss.

This dual-branch training strategy, which employs \(\mathcal{E}_\text{col}\) and \(\mathcal{E}_\text{geo}\) to encode the amplitude and phase from a colored point cloud, allows the model to process color and geometric attributes independently. Furthermore, based on the spectral convolution theorem~\cite{ffc_evidence} in Fourier theory, the operations performed by the encoders inherently affect the entire point cloud, leading to a large receptive field with a global influence. Additional details on the model's architecture and implementation are in the supplementary.

\subsection{Point Cloud Segmentation}
\label{subsec:segmentation}
To validate the scalability of our approach, we construct segmentation models for both part-level and large-scale scene-level tasks (\cref{fig:feature_extractor}(b)). For part segmentation, we employ a PointNet$++$~\cite{pointnet++}, while for outdoor scene segmentation, we utilize PointMetaBase-L~\cite{pointmeta} as the backbone. 
In both architectures, we integrate our proposed encoding by concatenating the feature representation $f_{\text{final}}$ with the intermediate features at each encoder stage. 
This simple integration allows the models to leverage explicit color and geometric attributes alongside the efficient feature extraction capabilities of the respective backbones.

\begin{figure*}[t]
    \centering
    \includegraphics[width=\linewidth]{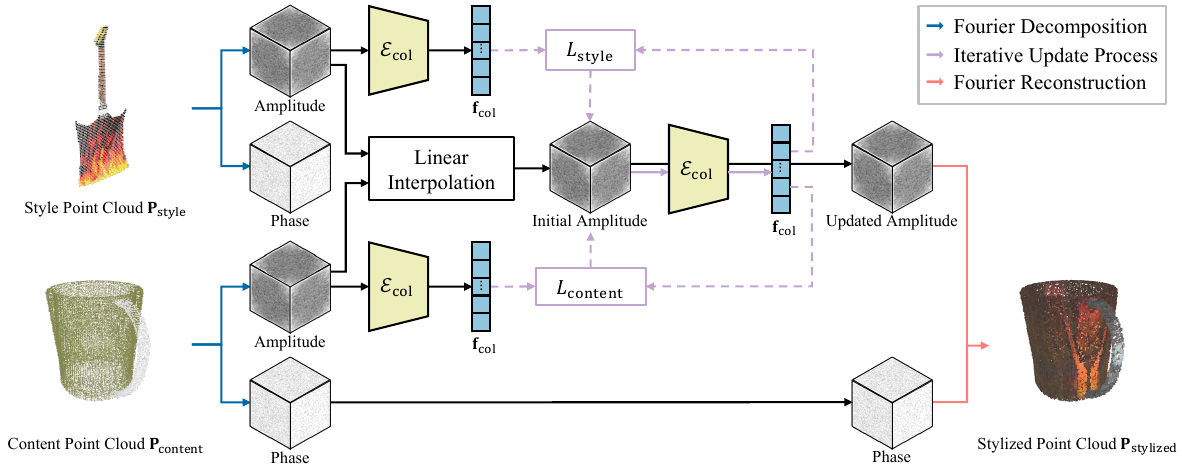}
    \caption{Overall pipeline of our encoding method applied to style transfer.
    The stylized point cloud's phase is initialized to be identical to that of the content point cloud, while its amplitude is initialized through a linear interpolation between the amplitudes of the content and style point clouds. Subsequently, features are extracted from the amplitudes of the style, content, and stylized point clouds using a color encoder $\mathcal{E}_\text{col}$. The loss is then computed, and the amplitude of the stylized point cloud is iteratively updated. Once the update process is complete, the final stylized point cloud is generated by performing Fourier reconstruction using the refined amplitude and phase.}
    \label{fig:style_transfer}
\end{figure*}

\subsection{Point Cloud Style Transfer}
\label{subsec:point_cloud_style_transfer}
We show the applicability of our encoding method by employing it in the \textit{style transfer} task.
While simply swapping amplitude allows us to transform a point cloud’s color (\textit{i.e.}, style) to resemble those of another, this approach struggled to achieve high-quality style transfer when the two point clouds had significantly different structures or colors. 
To enable style transfer between more disparate point clouds, we propose a simple pipeline that combines Fourier decomposition with optimization-based style transfer~\cite{psnet, style_transfer}.

\myparagraph{Overall Style Transfer Pipeline.}
Given a content point cloud $\mathbf{P}_\text{content}$ and a style point cloud $\mathbf{P}_\text{style}$, the goal of this task is to transfer the style (\textit{i.e.}, color attribute) of $\mathbf{P}_\text{style}$ onto $\mathbf{P}_\text{content}$. 
As shown in \cref{subsec:experimental_analysis}, Fourier decomposition separates a point cloud into amplitude and phase, where the amplitude encodes color attributes, and the phase captures geometric information.
Based on this analysis, we design the style transfer pipeline as illustrated in \cref{fig:style_transfer}. The process of style transfer is as follows: (1) Fourier decomposition is applied to both the content and style point clouds, yielding amplitude \(\mathcal{A}_\text{content}\) and phase \(\mathcal{P}_\text{content}\) from \(\mathbf{P}_\text{content}\), and amplitude \(\mathcal{A}_\text{style}\) and phase \(\mathcal{P}_\text{style}\) from \(\mathbf{P}_\text{style}\). 
(2) The amplitude $\mathcal{A}_\text{stylized}$ and phase $\mathcal{P}_\text{stylized}$ of the output point cloud (\textit{i.e.}, the stylized point cloud) is initialized by setting $\mathcal{P}_\text{stylized}$ to the phase of \(\mathbf{P}_\text{content}\), and initializing $\mathcal{A}_\text{stylized}$ through a linear interpolation of \(\mathcal{A}_\text{content}\) and \(\mathcal{A}_\text{style}\) as
\begin{equation}
\small
    \mathcal{A}_\text{stylized} = (1-\gamma) \mathcal{A}_\text{content} + \gamma \mathcal{A}_\text{style}.
\end{equation}
Here, $\gamma$ balances the ratio of the amplitudes, determining the intensity of the stylization effect.
Utilizing these features, we iteratively update the $\mathcal{A}_\text{stylized}$ according to the proposed loss formulation, optimizing it into the ideal amplitude of stylized point cloud.
Ultimately, the updated amplitude $\mathcal{A}_\text{stylized}$ undergoes Fourier reconstruction alongside the phase $\mathcal{P}_\text{stylized}$ to form the point cloud.

\myparagraph{Loss Formulation.}
Our pipeline updates the amplitude of the stylized point cloud through amplitude loss \(L_\text{amp}\). 
This loss encourages the amplitude of the stylized point cloud to incorporate the color attributes of both the content and style point clouds.
Accordingly, we compute $L_{\text{amp}}$ by combining the two loss terms, $L_{\text{content}}$ and $L_{\text{style}}$, with a weight $\alpha$, yielding \(L_\text{amp} = L_\text{content} + \alpha L_\text{style}\). 
The loss terms $L_\text{content}$ and $L_\text{style}$ measure the discrepancy between the amplitudes of the stylized point cloud and those of the content and style point clouds, respectively, and are defined as
\begin{equation}
\small
    L_\text{content} = \sum_{l\in\{2, 4, 7, 8\}} \bigl\lVert \mathcal{E}_\text{col}^l(\mathcal{A}_\text{content}) - \mathcal{E}_\text{col}^l(\mathcal{A}_\text{stylized}) \bigr\rVert^2,
\end{equation}
\begin{equation}
\small
    L_\text{style} = \sum_{l\in\{2, 4, 7, 8\}} \bigl\lVert \mathcal{E}_\text{col}^l(\mathcal{A}_\text{style}) - \mathcal{E}_\text{col}^l(\mathcal{A}_\text{stylized}) \bigr\rVert^2.
\end{equation}
Here, \(\mathcal{E}_\text{col}^l\) denotes the feature maps taken from the \(l\)-th layer of the color encoder. 
Through this formulation, the amplitude of the stylized point cloud is iteratively refined to semantically align with the amplitude features of both the content and style point clouds.

\myparagraph{Style Transfer from Image.}
This pipeline extends beyond style transfer between two point clouds and enables the transfer of the style of an image onto a point cloud. Since images are represented on a 2D pixel grid, adapting them to our style transfer framework requires transformation into a format analogous to the voxel grid obtained from the Fourier decomposition of a point cloud. To ensure compatibility with \(\mathcal{E}_\text{col}\), the image is resized to match the width \(W\) and height \(H\) determined during the voxelization of the point cloud. The resulting 2D array is then stacked along the depth dimension \(D\), forming a structure similar to a voxel grid. Fourier decomposition is subsequently applied to extract the amplitude, which is used as \(\mathcal{A}_\text{style}\) within the pipeline, facilitating the transfer of the image’s style onto the content point cloud.

\section{Experiments}
\label{sec:experiments}

In this section, we first describe the experimental settings, including implementation details and datasets (Sec.~\ref{exp:setup}). 
Then, we discuss the experiments for classification, segmentation (Sec.~\ref{exp:classification_segmentation}), and style transfer (Sec.~\ref{exp:style}), followed by their results and analysis. 
See the supplementary material for additional details.

\subsection{Experimental Setup}
\label{exp:setup}
\myparagraph{Dataset.} 
We evaluate the performance of our point cloud encoding methodology across various tasks using the DensePoint dataset~\cite{densepoint}, which is an extension of ShapeNet~\cite{shapenet} and ShapeNetPart~\cite{shapenetpart}. 
DensePoint consists of 16 classes and is \textit{the only dataset} specifically suited for colored object point cloud, comprising a total of 10,454 colored point clouds. 
We adhere to the official train-test split of DensePoint. 
However, since the number of point clouds per class in the training set is imbalanced, we randomly select 300 samples for classes with a larger number of instances, such as tables and chairs, to ensure a more balanced distribution. 
Furthermore, to evaluate the performance in outdoor scene segmentation, we additionally conduct experiments on the Semantic3D reduced-8 dataset~\cite{semantic3d}. 
Due to the large-scale nature of these point clouds, we employ a random cropping strategy where the data is sampled into cubic volumes with a side length of 7m.
All point clouds are normalized to fit within the range of \([-1, 1]\).
Although commonly used in point cloud research, datasets such as ModelNet40~\cite{modelnet}, which contain only uncolored point clouds, were not employed in our evaluation, as they are unsuitable for assessing the performance of our method specifically designed for encoding colored point clouds.

\myparagraph{Implementation Details.}
We utilize a single NVIDIA RTX A6000 for all experiments. 
In our voxelization process for Fourier decomposition, a voxel size of $v = 0.01$ was employed.
For classification and segmentation task, we train a model for 100 epochs with a batch size of 64, using an initial learning rate of 0.001 and the Adam optimizer. 
For the style transfer task, the amplitude of the stylized point cloud is updated over 5,000 iterations with an initial learning rate of 0.01 and is optimized using the Adam optimizer. For the amplitude loss ($L_\text{amp}$) computation, we set the weighting factor \(\alpha\) to 10. In the initialization process of the amplitude of the stylized point cloud, we set \(\gamma\) to 1, ensuring that the style of the style point cloud is maximally reflected.

\begin{table}[t]

    \begin{minipage}[t]{0.63\linewidth}
        \caption{Quantitative comparisons of our model against other methods in classification and part segmentation.}
        \vspace{-5pt}
        \centering
        \resizebox{\textwidth}{!}{    
        \begin{tabular}{l c c c c}
        \toprule
        \multirow{2}{*}{\centering Method} & \multicolumn{2}{c}{Classification} & \multicolumn{2}{c}{Part Segmentation} \\
        
        \cmidrule(lr){2-3} \cmidrule(lr){4-5}
         
        & {\centering OA ($\uparrow$)} & {\centering mAcc ($\uparrow$)} &  {\centering Cls. mIoU ($\uparrow$)} &
        {\centering Ins. mIoU ($\uparrow$)} \\
    
        \midrule
    
        PointNet~\cite{pointnet} & 97.25 & 95.82 & 81.84 & 84.11  \\
        PointNet$++$~\cite{pointnet++} & 97.54 & 97.28 & 82.06 & 84.26  \\
        PointConv~\cite{pointconv} & 97.54 & 96.01 & 83.13 & 84.90  \\
        DGCNN~\cite{dgcnn} & 98.26 & 97.56 & 82.27 & 84.54  \\
        PointTransformer~\cite{point_transformer} & 97.36 & 96.43 & 83.85 & 85.38  \\
        CurveNet~\cite{curvenet} & 97.40 & 96.48 & 83.46 & 85.79  \\
        PointMLP~\cite{pointmlp} & 97.97 & 97.85 & 83.43 & 85.94  \\
        PointNeXt~\cite{pointnext} & 98.13 & 97.72 & 83.91 & 86.16  \\
        PointVector~\cite{pointvector} & 97.81 & 96.27 & 84.63 & 86.80  \\
        PointMeta~\cite{pointmeta} & 98.11 & 97.58 & 84.87 & 87.61  \\
        Interpretable3D~\cite{interpretable3d} & 97.87 & 96.53 & 85.44 & 87.31  \\
        DeepLA~\cite{deepla} & 98.21 & 97.85 & 85.89 & 88.08  \\
    
        \midrule
        \rowcolor{blue!3}
        Ours & \textbf{98.43} & \textbf{97.92} & \textbf{86.03} & \textbf{88.21}  \\
        
        \bottomrule
        \end{tabular}
        }
        \label{tab:classification}
    \end{minipage}
    \hfill
    \begin{minipage}[t]{0.33\textwidth}
        \caption{Quantitative results on the Semantic3D (reduced-8) dataset for outdoor scene segmentation.}
        \vspace{-5pt}
        \centering
        \resizebox{\textwidth}{!}{    
        \begin{tabular}{l c}
            \toprule
            Method & mIoU ($\uparrow$) \\
            \midrule
            SegCloud~\cite{segcloud} & 61.3 \\
            RF\_MSSF~\cite{rf_mssf} & 62.7 \\
            SPG~\cite{spg} & 73.2 \\
            ShellNet~\cite{shellnet} & 69.4 \\
            GACNet~\cite{gacnet} & 70.8 \\
            FGCN~\cite{fgcn} & 62.4 \\
            PointGCR~\cite{pointgcr} & 69.5 \\
            RandLA~\cite{randla} & 77.4 \\
            KPConv rigid~\cite{kpconv} & 74.6 \\
            KPConv deform~\cite{kpconv} & 73.1 \\
            RFCR~\cite{rfcr} & 77.8 \\
            \rowcolor{blue!3}
            \midrule
            Ours & \textbf{78.0} \\
            \bottomrule
        \end{tabular}
        }
        \label{tab:semantic3d}
    \end{minipage}
    \vspace{-10pt}
\end{table}

\subsection{Classification and Segmentation}
\label{exp:classification_segmentation}
\myparagraph{Metrics and Baselines.} 
To evaluate the performance of our classification and segmentation models incorporating our encoding approach (described in \cref{subsec:classification,subsec:segmentation}), we employ overall accuracy (OA, \%), mean per-class accuracy (mAcc, \%), 
and mean intersection over union (mIoU, \%) in both class-wise (Cls.) and instance-wise (Ins.) forms as evaluation metrics.
As there exists \textit{no prior work} proposing an encoding approach specifically tailored for colored point clouds like ours, we compare our method against widely used point cloud processing models that commonly serve as backbone architectures across various tasks.

\begin{table}[t]

    \begin{minipage}[t]{0.48\linewidth}
        \vspace{0pt}
        \centering
        \includegraphics[width=\linewidth]{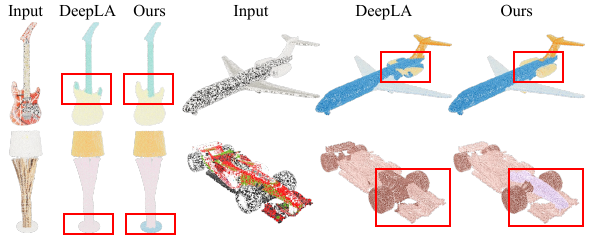}
        \vspace{15pt}
        \captionof{figure}{
        Qualitative part segmentation results on DensePoint~\cite{densepoint}. 
        The regions marked with red squares indicate areas of significant performance improvement.
        }
        \label{fig:segmentation}
    \end{minipage}
    \hfill
    \begin{minipage}[t]{0.48\textwidth}
        \vspace{0pt}
        \centering
        \includegraphics[width=\linewidth]{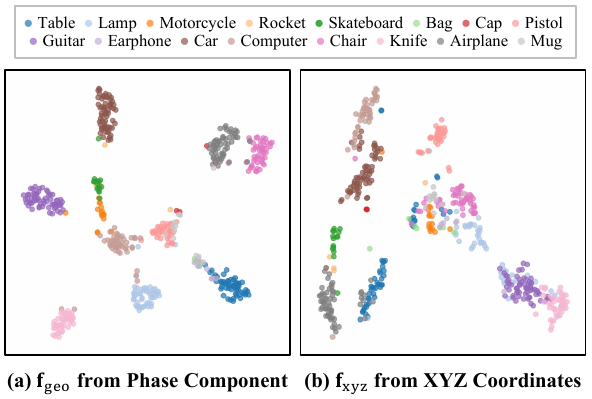}
        \captionof{figure}{
        t-SNE visualization of features extracted from the phase component (\(\mathbf{f}_\text{geo}\)) and XYZ coordinates (\(\mathbf{f}_\text{xyz}\)).
        }
        \label{fig:tsne}
    \end{minipage}
    \vspace{-15pt}
\end{table}

\myparagraph{Evaluation Results.}
As shown in \Cref{tab:classification,tab:semantic3d}, our classification and segmentation models performs favorably against other point cloud processing methods.
Our model achieved the highest OA of 98.43\% and the highest mAcc of 97.92\% in classification, while demonstrating superior performance in part segmentation with a class mIoU of 86.03\% and an instance mIoU of 88.21\%.
In particular, for large-scale outdoor scene segmentation, our method achieves a state-of-the-art mIoU of 78.0\% on the Semantic3D dataset.
As shown in \cref{fig:segmentation}, the qualitative results 
also demonstrate the superiority of our approach.
These results demonstrate the superior effectiveness of our colored point cloud encoding methodology compared to existing approaches.
Moreover, unlike conventional voxelization-based approaches~\cite{voxnet, voxelization2} that incur significant memory costs, our classification model effectively utilizes voxelization while maintaining a remarkably lower number of parameters compared to other state-of-the-art methods.
For instance, while models such as PointMLP~\cite{pointmlp} and PointNeXt~\cite{pointnext} require 13.2M and 46.1M parameters respectively, our model achieves superior performance with only 3.5M parameters, which is more than 13 times fewer than PointNeXt.

\myparagraph{Analysis Through t-SNE Visualization.}
To verify that our approach encodes geometric attributes more effectively than split-based point cloud encoding approach, we compared the feature \(\mathbf{f}_{\text{xyz}}\), generated by processing raw XYZ coordinates into a representative feature encoder~\cite{pointnet}, with \(\mathbf{f}_{\text{geo}}\) using t-SNE visualization~\cite{tsne}.
When visualizing \(\mathbf{f}_{\text{geo}}\) (\cref{fig:tsne}(a)), we observe that point clouds belonging to the same class tend to form distinct clusters, demonstrating that \(\mathbf{f}_{\text{geo}}\) effectively captures class-relevant geometric features.
In contrast, the visualization of \(\mathbf{f}_{\text{xyz}}\) (\cref{fig:tsne}(b)) reveals significant overlap between point clouds of different classes, indicating less distinct separation among them.
This implies that \(\mathbf{f}_{\text{geo}}\) incorporates the semantic information essential for determining the class of a 3D object more effectively than \(\mathbf{f}_{\text{xyz}}\).

\subsection{Point Cloud Style Transfer}
\label{exp:style}
As there are \textit{no standardized metrics} for evaluating point cloud style transfer, we assess the effectiveness of our method through qualitative analysis. 
Specifically, we evaluate how effectively the color attributes of a style point cloud or image are transferred to different content point clouds.

\begin{table}[t]

    \begin{minipage}[t]{0.48\linewidth}
        \centering
        \includegraphics[width=\linewidth]{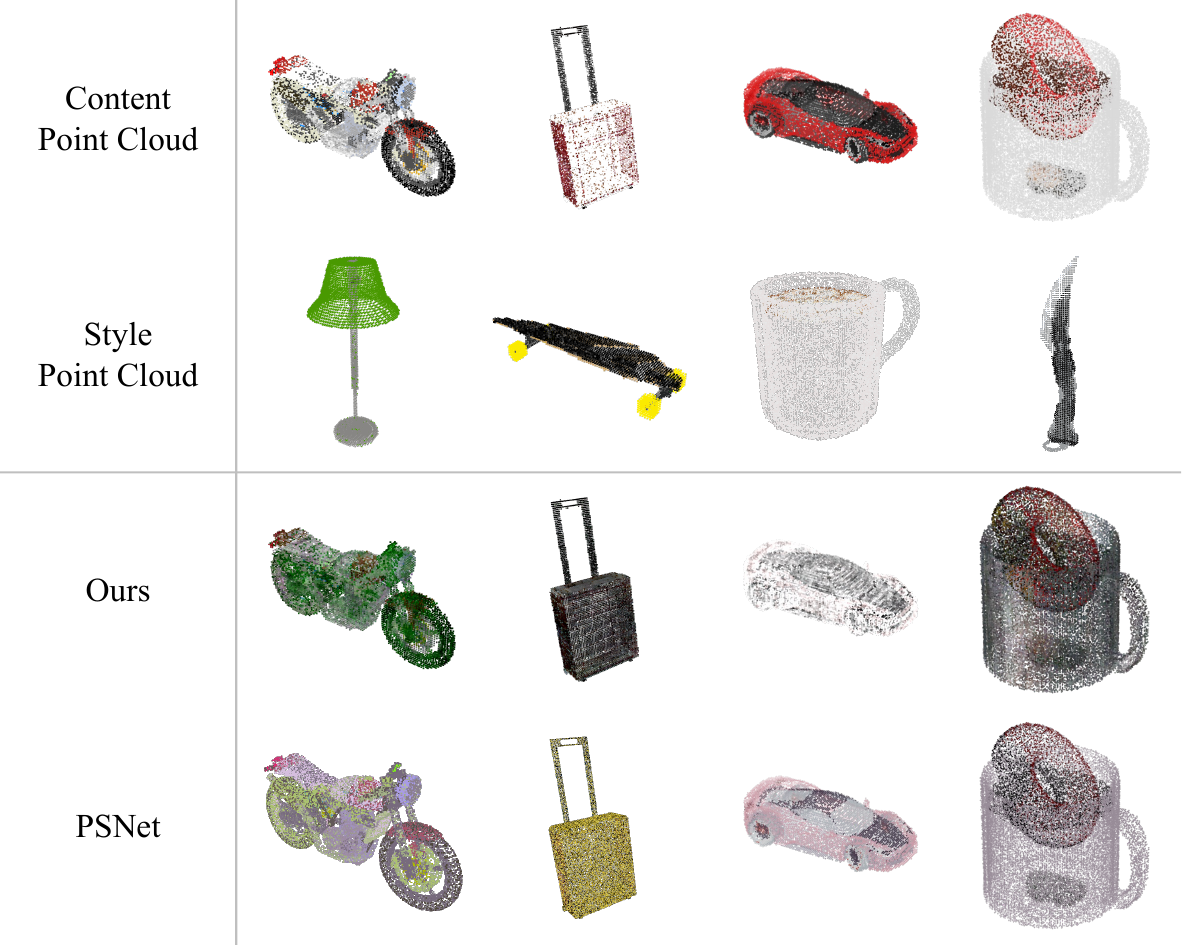}
        \captionof{figure}{Qualitative comparison of point cloud-to-point cloud style transfer results between PSNet~\cite{psnet} and our method.}
        \label{fig:supp_pcd_pcd}
    \end{minipage}
    \hfill
    \begin{minipage}[t]{0.48\textwidth}
        \centering
        \includegraphics[width=\linewidth]{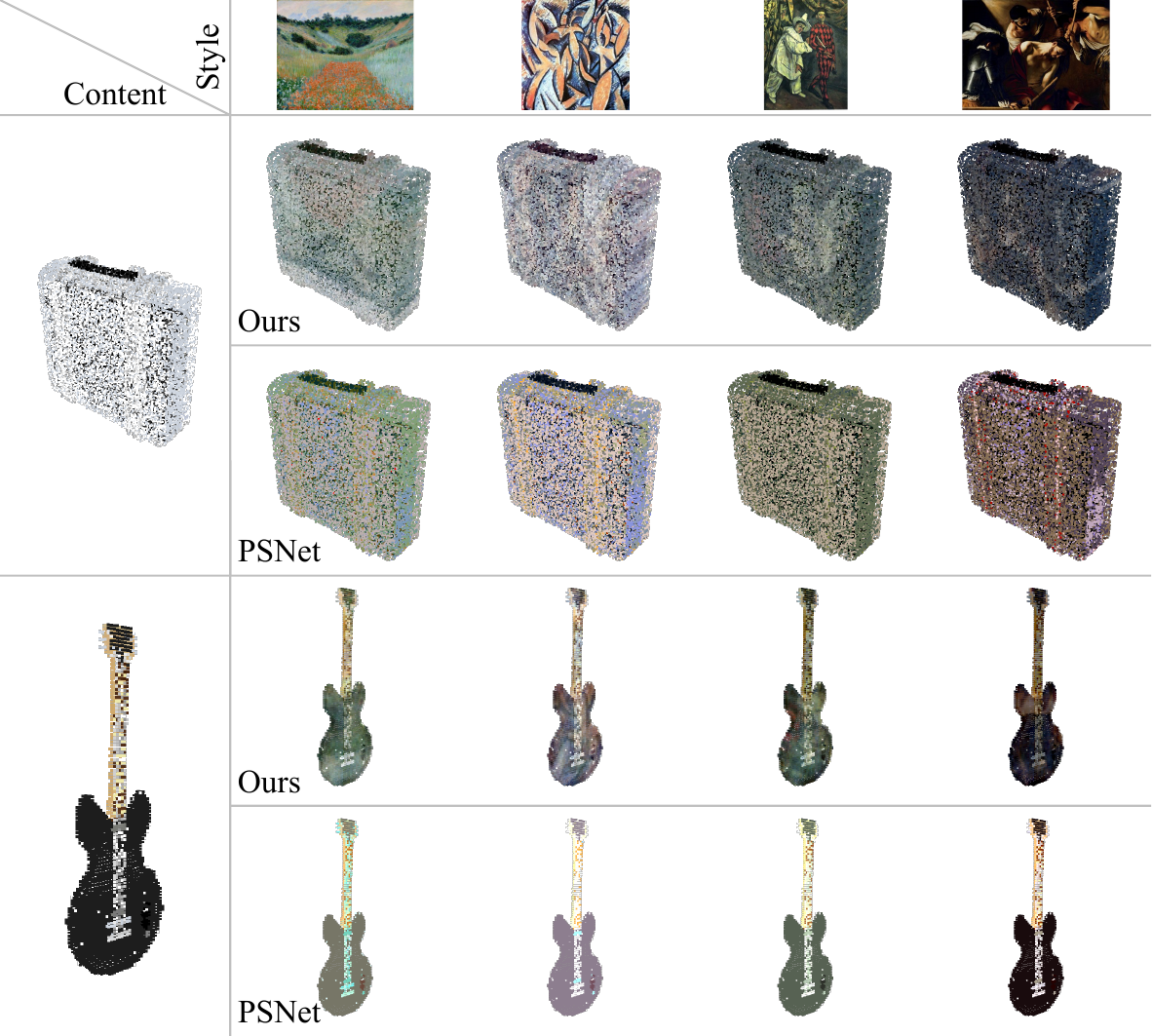}
        \captionof{figure}{Qualitative comparison of image-to-point cloud style transfer results between PSNet~\cite{psnet} and our method.}
        \label{fig:supp_pcd_img}
    \end{minipage}
    \vspace{-15pt}
\end{table}

\myparagraph{Qualitative Results.}
We compare our method against PSNet~\cite{psnet}, which stands as the sole established model specifically designed for colored point cloud style transfer. 
PSNet optimizes both point positions and colors in the point cloud through a dual-encoder framework, but its limited receptive field and lack of consideration for neighboring points constrain its ability to capture and transfer the global contextual color patterns of the style images.
In contrast, our approach utilizes phase and amplitude information, enabling style transfer with an expanded receptive field. 
Consequently, as shown in \cref{fig:supp_pcd_pcd,fig:supp_pcd_img}, our method preserves the global color patterns of the style image or point cloud in the transferred point cloud. 
This comparison shows the advantage of using Fourier decomposition for explicit disentanglement of color and geometric attributes, resulting in improved stylization quality in colored point clouds. 
Additional style transfer results are in the supplementary.

\subsection{Additional Analyses}

\begin{table}[t]

    \begin{minipage}[t]{0.33\linewidth}
        \vspace{0pt}
        \centering
        \caption{Ablation study on different encoding strategies using PointMeta~\cite{pointmeta} as the backbone. Our Fourier-based encoding significantly improves classification accuracy.}
        \resizebox{\textwidth}{!}{  
        \begin{tabular}{l c c}
            \toprule
            Encoding & OA ($\uparrow$) & mAcc ($\uparrow$) \\
            \midrule
            Traditional & 98.11 & 97.58 \\
            Split-based & 98.19 & 97.71 \\
            \rowcolor{blue!3}
            {Ours} & \textbf{98.43} & \textbf{97.92} \\
            \bottomrule
        \end{tabular}
        }
        \label{tab:ablation_decomposition}
    \end{minipage}
    \hfill
    \begin{minipage}[t]{0.63\textwidth}
        \vspace{0pt}
        \centering
        \includegraphics[width=0.95\linewidth]{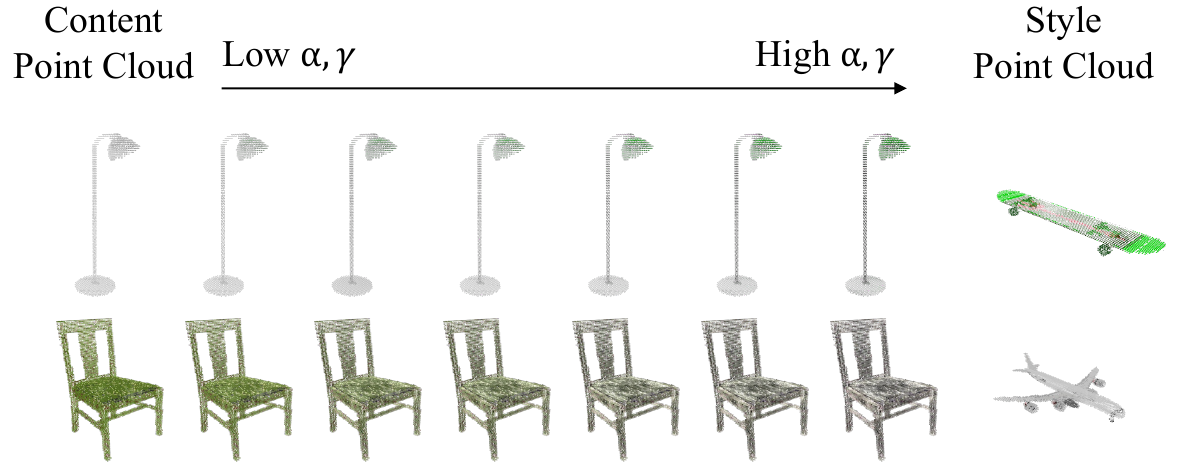}
        \captionof{figure}{Experimental analysis of hyperparameters used in loss calculation and stylized point cloud amplitude initialization. 
        We evaluate the impact of $\alpha$ and $\gamma$ on the stylization quality. Increasing both parameters leads to a more pronounced transfer of color attributes from the style point cloud to the content point cloud.
        }
        \label{fig:ablation1}
    \end{minipage}
    \vspace{-15pt}
\end{table}

\myparagraph{Ablation Study on Encoding Strategies.}
To verify the impact of our Fourier-based decomposition, we conduct an ablation study comparing different encoding strategies using PointMeta as the common backbone. 
As shown in \Cref{tab:ablation_decomposition}, the traditional (\cref{fig:overview_2col}(a)) and the split-based approach (\cref{fig:overview_2col}(b)) show lower performance compared to our method (\cref{fig:overview_2col}(c)). 
Our Fourier-based encoding achieves the highest performance across all metrics by effectively disentangling attributes in the spectral domain, which inherently provides a larger receptive field and global contextual awareness.

\myparagraph{Hyperparameter in Style Transfer Pipeline.}
\cref{fig:ablation1} illustrates how the final style transfer outcome varies depending on the values of \( \alpha \) and \( \gamma \). 
Since both $\alpha$ and $\gamma$ serve the same purpose of controlling the degree of style variation, we varied both simultaneously to compare the resulting outputs.
When both parameters are set to small values, the resulting stylized point cloud exhibits minimal influence from the color attributes of the style point cloud. Conversely, larger values of \( \alpha \) and \( \gamma \) lead to a significant stylistic transformation, with the stylized point cloud closely mirroring the color attributes of the style point cloud.

\begin{figure*}[t]
    \centering
    \includegraphics[width=0.6\linewidth]{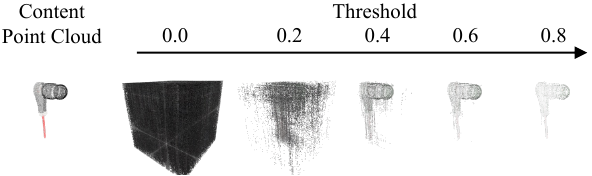}
    \caption{
    Style transfer results with \(\pi\)-value-based noise removal.
    Voxels with \(\pi\) below a predefined threshold are removed as noise. A low threshold leaves residual noise, while increasing it gradually reduces noise. 
    However, an excessively high threshold removes essential points, leading to information loss.
    }
    \label{fig:ablation2}
    \vspace{-10pt}
\end{figure*}

\myparagraph{Threshold for Noise Removal Using $\pi$ Value.}
\cref{fig:ablation2} illustrates the impact of different thresholds on style transfer results. When the threshold is set to 0.0, meaning no noise removal is applied, the resulting point cloud is overwhelmingly obscured by noise. However, as the threshold increases, noise is progressively eliminated, yielding a more refined point cloud. Nevertheless, an excessively high threshold may erroneously classify essential points as noise, leading to their removal. Based on this observation, we set the threshold to 0.5 in our experiments.
\section{Conclusion}
In this paper, we introduce a colored point cloud encoding method leveraging 3D Fourier decomposition to explicitly represent color and geometric attributes. By decomposing point clouds into amplitude and phase, our approach independently captures color (via amplitude) and geometry (via phase), while benefiting from a wide receptive field ensured by the spectral convolution theorem.
Our method improves classification, segmentation, and style transfer, achieving SOTA performance.
These findings show our approach's effectiveness in capturing color and geometry, laying the foundation for future 3D vision research.

\bibliographystyle{splncs04}
\bibliography{main}

\clearpage

\setcounter{section}{0} 
\renewcommand{\thesection}{\Alph{section}}
\renewcommand{\thesubsection}{\thesection.\arabic{subsection}}

\section{Theoretical Rationale for Inherent Property of Amplitude and Phase}
In the realm of three-dimensional signal analysis, the discrete Fourier transform (DFT) serves as an instrument that decomposes a 3D data into its fundamental frequency components. 
The transformation is formally represented as
\begin{equation}
    F(k, l, m) = \sum_{x=0}^{N-1} \sum_{y=0}^{M-1} \sum_{z=0}^{L-1} f(x, y, z) \exp\!\Bigl(-2\pi i\Bigl(\frac{kx}{N} + \frac{ly}{M} + \frac{mz}{L}\Bigr)\Bigr).
\end{equation}
Here, \( f(x, y, z) \) represents the original data distributed over a discrete three-dimensional voxel grid, while \( F(k, l, m) \) denotes its frequency-domain counterpart. The indices \( k \), \( l \), and \( m \) traverse the frequency space corresponding to the spatial dimensions \( x \), \( y \), and \( z \) respectively.

Each complex coefficient \( F(k, l, m) \) inherently embodies two vital aspects: its amplitude and its phase. The amplitude, defined as
\begin{equation}
    \text{Amplitude} = |F(k, l, m)|,
\end{equation}
represents the magnitude of the complex number, conveying the intensity or \textit{vividness} of a particular frequency component. 
Conceptually, the amplitude of each frequency component is analogous to the \textit{color attributes} of 3D data; it quantifies the intensity and effectively represents the brightness distribution within the multidimensional structure.

Conversely, the phase is determined by the argument of the complex number as
\begin{equation}
    \text{Phase} = \arg\bigl(F(k, l, m)\bigr),
\end{equation}
and it captures the intrinsic geometric configuration of the data. The phase orchestrates the spatial arrangement and intricate morphology of the original array, encoding the positional information. 
In this regard, while the amplitude outlines the brightness or color, the phase is the component, representing the geometry and form of the three-dimensional structure.

Thus, the 3D discrete Fourier transform not only transforms the spatial data into a frequency spectrum but also decomposes it into two symbiotic realms: one that embodies the color attribute and another that encapsulates the geometric attribute. This duality offers a profound insight into the underlying characteristics of the data, marrying the realms of vivid intensity with structural geometry.

\section{Architecture Details}
We detail the architecture, which consists of three main components: (\expandafter{\romannumeral1}) Fourier decomposition, (\expandafter{\romannumeral2}) a feature extractor, and (\expandafter{\romannumeral3}) a fusion module. Our approach utilizes Fourier decomposition to explicitly separate color information from geometric structure (\ie, extracting amplitude that captures fine color details and phase that encodes intrinsic geometry). This explicit disentanglement enables dedicated processing modules for color and geometry, ultimately yielding a robust, discriminative feature representation that enhances classification and segmentation performance on colored point cloud. Although the main paper already provides a detailed description of the Fourier decomposition, the following supplementary sections offer a comprehensive explanation of (\expandafter{\romannumeral2}) the feature extractor and (\expandafter{\romannumeral3}) the fusion module.

\begin{figure*}[t]
    \centering
    \includegraphics[width=0.7\linewidth]{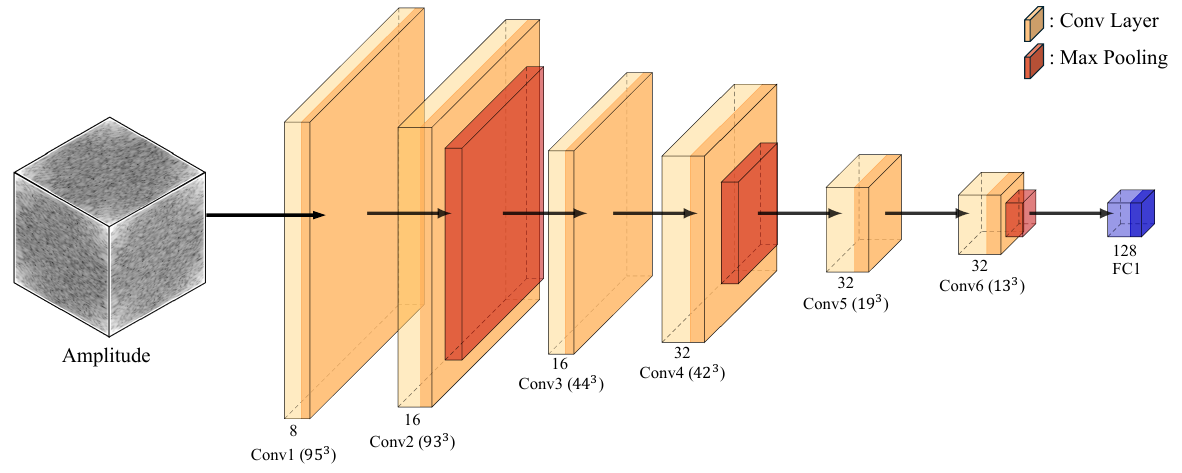}

    \caption{Detailed illustration of \( \mathcal{E}_\text{col} \) introduced in Sec~\ref{supp:feature_extractor}. Our encoder, consisting of multiple layers of 3D convolutions, processes frequency components as input and outputs a one-dimensional feature vector. The geometry encoder \( \mathcal{E}_\text{geo} \), which takes the phase as input, follows the same architectural structure.}
    \label{fig:supp_feature_extractor}
\end{figure*}

\subsection{Feature Extractor}
\label{supp:feature_extractor}
Our feature extractor employs identical 3D convolutional encoders separately for amplitude $\mathcal{A}$ and phase representation $\mathcal{P}$, where the phase encoder is denoted as \( \mathcal{E}_\text{geo} \) and the amplitude encoder as \( \mathcal{E}_\text{col} \).
As illustrated in Fig.~\ref{fig:supp_feature_extractor}, each encoder comprises six 3D convolutional layers with kernel sizes of 5 or 7, interleaved with Leaky ReLU activations (\(\alpha = 0.1\)).
Three 3D max-pooling layers are interspersed among the convolutions to progressively reduce spatial resolution and aggregate multi-scale features.
Finally, a fully connected layer flattens the last convolutional feature map and outputs a 128-dimensional feature vector.
This encoding process is formally expressed as
\begin{equation}
    \mathbf{f}_{\text{col}} = \mathcal{E}_{\text{col}}(\mathcal{A}), 
    \quad
    \mathbf{f}_{\text{geo}} = \mathcal{E}_{\text{geo}}(\mathcal{P})  .  
    \label{temp1}
\end{equation}
Here, \(\mathbf{f}_{\text{col}}\) and \(\mathbf{f}_{\text{geo}}\) denote the feature vectors that capture the color and geometric information of the point cloud, respectively.

\subsection{Fusion Module}
Given the color and geometric feature representations \( \mathbf{f}_{\text{col}} \) and \( \mathbf{f}_{\text{geo}} \) from Eq.~\eqref{temp1}, we concatenate them along the feature dimension to form the intermediate representation \( \mathbf{f}_{\text{concat}} \) as
\begin{equation}
    \mathbf{f}_{\text{concat}} = \left[ \mathbf{f}_{\text{col}}; \mathbf{f}_{\text{geo}} \right], 
    \label{temp2}
\end{equation}
resulting in a 256-dimensional representation that integrates the color and geometric cues of the point cloud. This concatenated feature $\mathbf{f}_{\text{concat}}$ is then refined by a fusion module, denoted as \(\mathcal{M}_{\mathrm{fus}}\), to yield the final feature representation $\mathbf{f}_{\text{final}}$ as
\begin{equation}
     \mathbf{f}_{\text{final}} = \mathcal{M}_{\mathrm{fus}}\left(\mathbf{f}_{\text{concat}}\right),  
     \label{temp3}
\end{equation}
where the fusion module \( \mathcal{M}_{\mathrm{fus}} \) is a two-layer fully connected network designed to refine and regularize the combined features through Batch Normalization and Dropout. Specifically, in the first layer, Batch Normalization is applied, followed by Dropout with a rate of 0.3. A linear transformation then maps the 256-dimensional input to 128 dimensions, followed by a ReLU activation. The second layer follows the same pattern, ultimately mapping the 128-dimensional features to the target class space (16 classes in DensePoint).  
Formally, the predicted logits \( \mathbf{z} \) are as
\begin{equation}
    \mathbf{z} = \mathrm{FC}_2\Big(\operatorname{Dropout}\big(\operatorname{BN}\big(\operatorname{ReLU}\big(\mathrm{FC}_1(\operatorname{Dropout}(\operatorname{BN}(\mathbf{f}_{\text{concat}})))\big)\big)\big)\Big),
\end{equation}
where FC\(_1\) and FC\(_2\) are the first and second fully connected layers of the fusion module \( \mathcal{M}_{\text{fus}} \), respectively.

\section{Additional Analyses}

\begin{table}[h]
    \centering
    \caption{Quantitative evaluation of color attribute transfer via KL-Divergence (N=1000). Lower values indicate higher similarity in color distribution. The results confirm that swapped point clouds closely follow their respective amplitude sources.}
    \begin{tabular}{l c c c}
        \toprule
        Comparison Pair & Relationship & Mean KL ($\downarrow$) & Std. Dev. \\
        \midrule
        $\text{KL}(P_1, P_1^{\text{AmpSwap}})$ & Geometry Source & 9.5657 & 3.8384 \\
        $\mathbf{KL}(P_2, P_1^{\text{AmpSwap}})$ & \textbf{Color Source} & \textbf{2.1284} & \textbf{0.5431} \\
        \midrule
        $\text{KL}(P_2, P_2^{\text{AmpSwap}})$ & Geometry Source & 5.2097 & 2.6846 \\
        $\mathbf{KL}(P_1, P_2^{\text{AmpSwap}})$ & \textbf{Color Source} & \textbf{2.2191} & \textbf{0.8250} \\
        \bottomrule
    \end{tabular}
    \label{tab:supp_kl_div}
\end{table}

\begin{figure}[h]
    \centering
    \includegraphics[width=\textwidth]{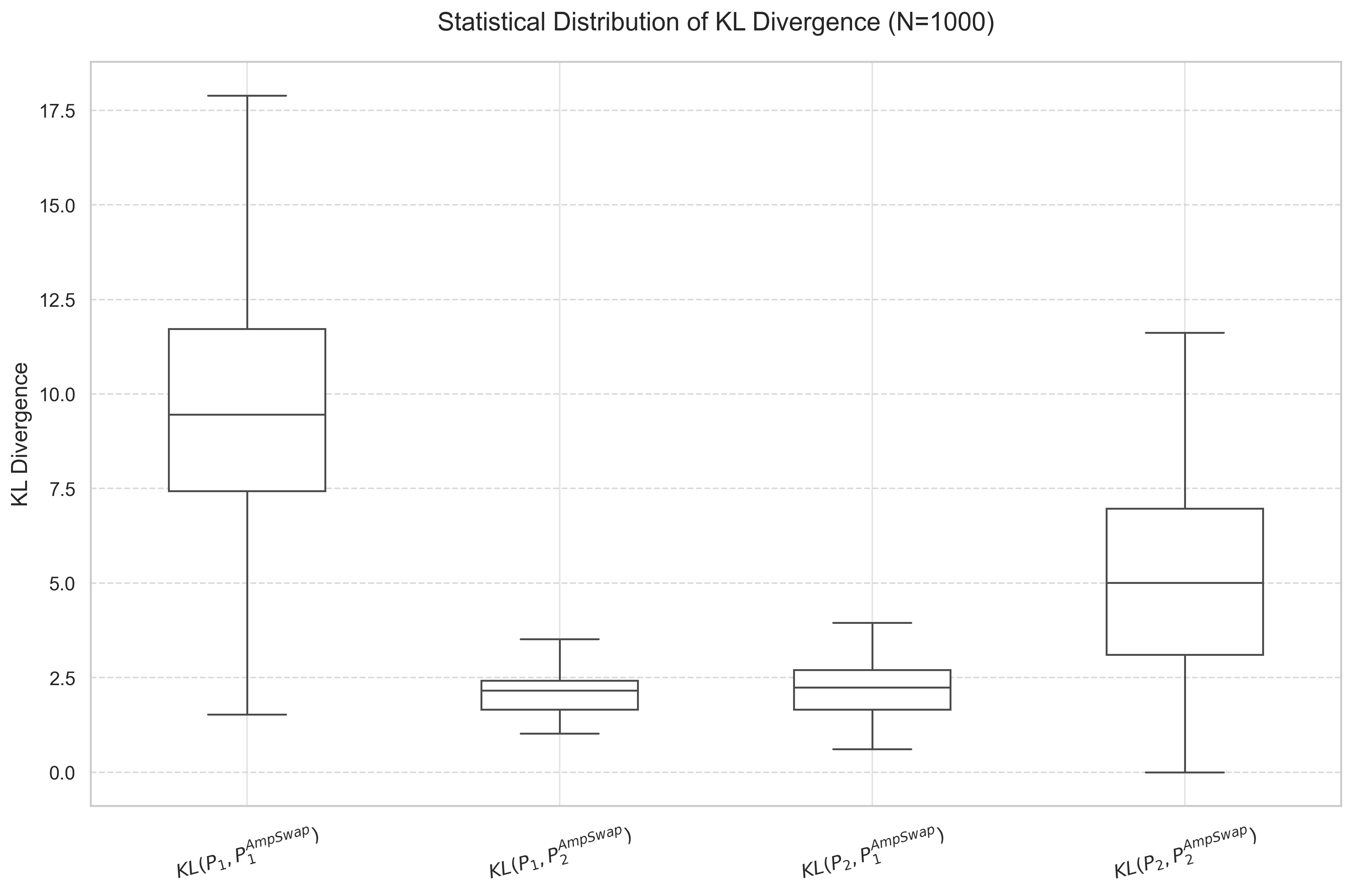}
    \caption{Statistical distribution of KL-Divergence over 1000 random trials. The significant gap between the target style source and the geometry source confirms the effective isolation of color attributes within the amplitude component.}
    \label{fig:supp_kl_dist}
\end{figure}

\subsection{Quantitative Analysis of Attribute Disentanglement}

To provide a more rigorous validation of our findings regarding attribute disentanglement, we conduct a quantitative analysis using KL-Divergence to measure the similarity between the color distributions of the original and swapped point clouds. While the visual results in Sec. 3.2 of the main paper provide intuitive evidence, this statistical approach confirms that the disentanglement is consistent across various samples.

\myparagraph{Experimental Setup.} 
We randomly sample 1000 pairs of colored point clouds from the DensePoint dataset~\cite{densepoint}. For each pair $(P_1, P_2)$, we perform the amplitude swap process as defined in Eq.~(4) of the main paper to generate $P_{i}^{\text{AmpSwap}}$. We then extract the RGB color histograms for each point cloud and calculate the average KL-Divergence across the three color channels.

\myparagraph{Results.} 
As summarized in \Cref{tab:supp_kl_div} and \cref{fig:supp_kl_dist}, the results consistently demonstrate that the color profile of the reconstructed point cloud follows its amplitude source. 
Specifically, the KL-Divergence between the swapped result $P_{1}^{\text{AmpSwap}}$ and its style source $P_2$ is significantly lower (mean: $2.1284$) compared to its divergence from the original geometric source $P_1$ (mean: $9.5657$). 
This bidirectional consistency—where $P_{2}^{\text{AmpSwap}}$ also accurately mirrors the color distribution of $P_1$ (mean KL: $2.2191$) while diverging from $P_2$ (mean KL: $5.2097$)—conclusively proves that the amplitude component explicitly represents the color attribute.

\begin{figure}[t]
    \centering
    \includegraphics[width=0.9\linewidth]{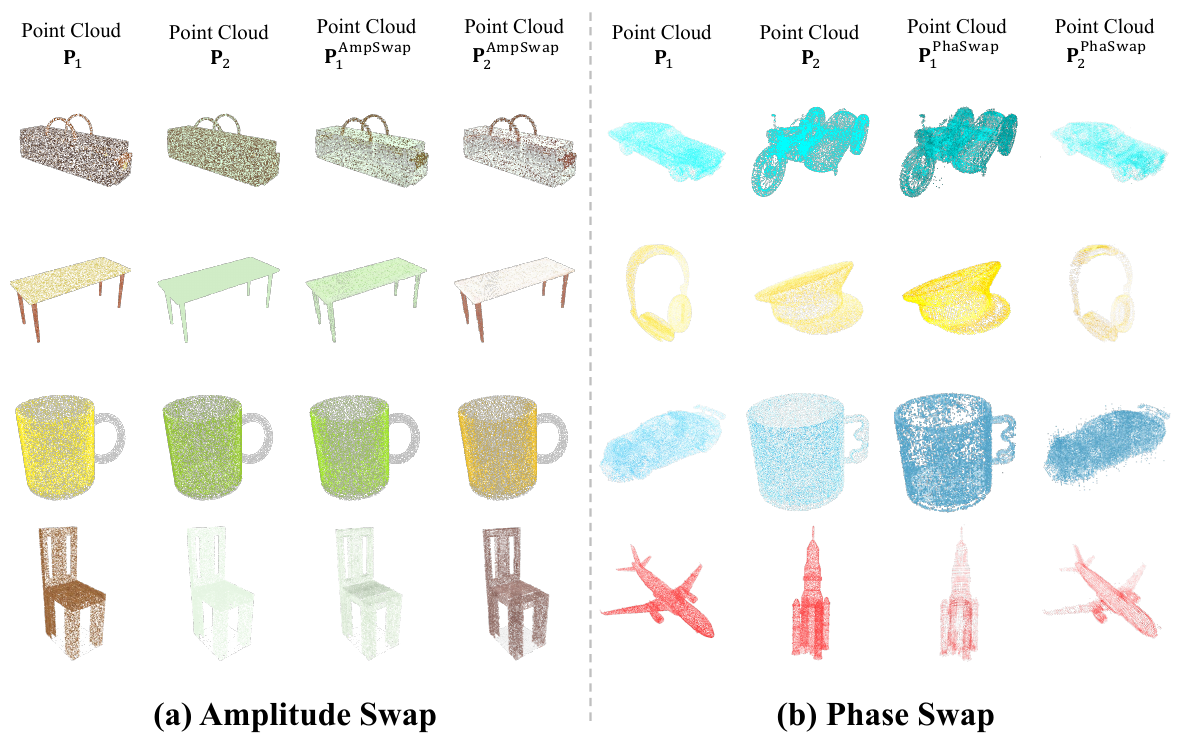}

    \caption{Visualization results of the amplitude swap and phase swap experiments.}
    \label{fig:supp_fourier_swap}
\end{figure}

\subsection{Fourier Swap: Additional Visual Results}
Additional results from the amplitude swap and phase swap experiments conducted in Sec. 3.2 of the main paper are presented in \cref{fig:supp_fourier_swap}. 
In \cref{fig:supp_fourier_swap}(a), the amplitude swap experiment demonstrates the exchange of color attributes between the two point clouds. 
Meanwhile, \cref{fig:supp_fourier_swap}(b) illustrates how swapping the phase components results in an interchange of their geometric structures.

\begin{figure}[t]
    \centering
    \begin{subfigure}[b]{0.48\textwidth}
        \centering
        \includegraphics[width=\textwidth]{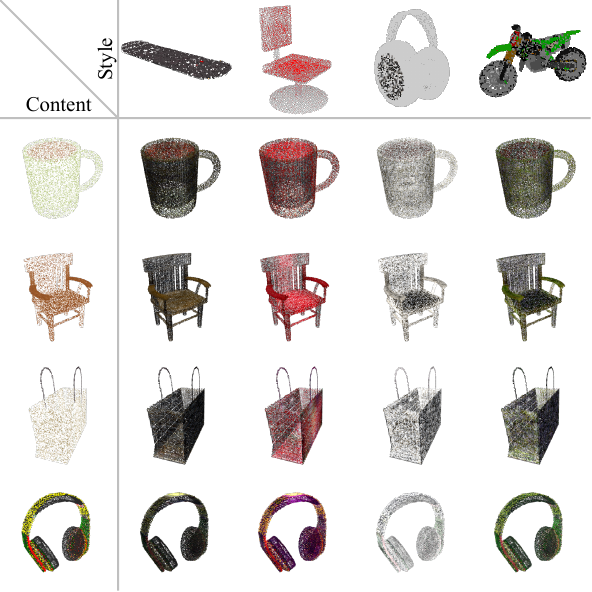}
        \label{fig:results_pcd_pcd}
    \end{subfigure}
    \hfill
    \begin{subfigure}[b]{0.48\textwidth}
        \centering
        \includegraphics[width=\textwidth]{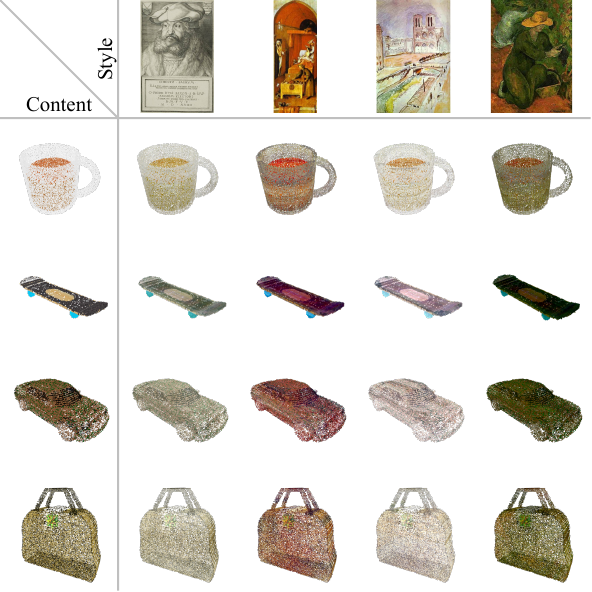}
        \label{fig:results_pcd_img}
    \end{subfigure}
    \caption{
    Visualization of point cloud style transfer results. Our method enables each content point cloud to adopt the color characteristics of the corresponding style point cloud or image while preserving its original geometric structure. This demonstrates the effectiveness of our approach in transferring style across diverse 3D objects while maintaining structural integrity.
    }
    \label{fig:main_style_transfer_results}
\end{figure}

\subsection{Style Transfer: Additional Visual Results}
As shown in \cref{fig:main_style_transfer_results}, the style (\textit{i.e.}, color attribute) of the style point cloud is effectively transferred to the content point cloud.
Due to the sparse nature of point clouds, it is challenging to capture fine-grained details as in images; however, the overall color tone and visual impression are well preserved.
Additionally, \cref{fig:main_style_transfer_results} demonstrates that our method successfully applies not only point cloud-to-point cloud style transfer but also image-to-point cloud style transfer, yielding promising results.

\section{Discussion}

\subsection{Limitations}
To encode an extensive scene, an enormous number of voxels is required during the voxelization process, leading to a dramatic increase in computational overhead.
Moreover, if the voxel size is not set sufficiently small during the voxelization process in Fourier decomposition, fine structural details of the point cloud may be lost.
Nevertheless, our point cloud encoding approach offers significant advantages over many existing point cloud encoding methods~\cite{pointnet, pointnet++, point_transformer, point_transformer_v2, point_transformer_v3, pointmlp, pointconv, kpconv}, as they achieve excellent performance while maintaining a low memory cost.

\subsection{Future Works}
Our point cloud encoding method not only enables the independent handling of color and geometric attributes but also leverages a large receptive field for feature extraction. 
As a result, while this paper does not demonstrate it due to the lack of a colored point cloud dataset, the approach can extend beyond classification, style transfer, and data augmentation to various point cloud tasks, such as generation and part segmentation.
Expanding the application of this encoding technique to a broader range of tasks remains a promising direction for future work.

\end{document}